\pdfoutput=1
\documentclass[11pt]{article}
\usepackage[]{acl}
\usepackage{times}
\usepackage{hyperref}
\usepackage{latexsym}
\usepackage{amsthm}
\usepackage{amsfonts}
\usepackage{graphicx}
\usepackage{multirow}
\usepackage{pbox}
\usepackage{float}
\usepackage{adjustbox}
\usepackage[T1]{fontenc}

\usepackage[utf8]{inputenc}

\usepackage{microtype}
\title{Biaffine Discourse Dependency Parsing}

\author{Yingxue Fu \\
  School of Computer Science\\ 
  University of St Andrews\\
  KY16 9SX,UK \\
  \texttt{yf30@st-andrews.ac.uk} \\}

\begin{document}
\maketitle
\begin{abstract}

We provide a study of using the biaffine model for neural discourse dependency parsing and achieve significant performance improvement compared with the baseline parsers. We compare the Eisner algorithm and the Chu-Liu-Edmonds algorithm in the task and find that using the Chu-Liu-Edmonds algorithm generates deeper trees and achieves better performance. We also evaluate the structure of the output of the parser with average maximum path length and average proportion of leaf nodes and find that the dependency trees generated by the parser are close to the gold trees. As the corpus allows non-projective structures, we analyze the complexity of non-projectivity of the corpus and find that the dependency structures in this corpus have gap degree at most one and edge degree at most one.

\end{abstract}

\section{Introduction}
Discourse parsing aims at uncovering the structure of argumentation and information flow of a text. It is associated with the coherence and naturalness of a text. Applications that may benefit from discourse information include text summarization~\citep{marcu-1997-discourse}, sentiment analysis~\citep{mukherjee-bhattacharyya-2012-sentiment}, essay scoring~\citep{nadeem-etal-2019-automated}, text classification~\citep{ji-smith-2017-neural}, machine translation~\citep{sim-smith-2017-integrating} and so on. 

Several frameworks for discourse parsing have been proposed. Rhetorical Structure Theory (RST)~\citep{mann1988rhetorical} assumes that discourse structure can be represented by a tree. It addresses text organization by means of relations that hold between discourse units, and a text is a hierarchically connected structure, in which each part has a role to play~\citep{taboada2006rhetorical}. RST discourse parsing resembles constituency-based syntactic parsing, and facilitated by the creation of the RST discourse corpus~\citep{carlson-etal-2001-building}, the RST framework is widely used in computational discourse processing. Another influential framework is the Penn Discourse Treebank (PDTB) style discourse parsing~\citep{prasad-etal-2008-penn, prasad-etal-2018-discourse}, which focuses on explicit and implicit local discourse relations. The Segmented Discourse Representation Theory (SDRT)~\citep{asher2003logics}  and the Graphbank~\citep{wolf-gibson-2005-representing} use graph to represent discourse structure.

~\citet{li-etal-2014-text} introduce the dependency framework into discourse parsing. Compared with RST discourse parsing, discourse dependency parsing is more flexible and potentially capable of capturing non-projective structures. 

Because of a lack of discourse corpora annotated under the dependency framework, previous studies center around finding automatic means of converting RST tree structures to discourse dependency structures. Studies in this direction include ~\citet{li-etal-2014-text},~\citet{hirao-etal-2013-single}, ~\citet{yoshida-etal-2014-dependency} and~\citet{morey-etal-2018-dependency}. These methods mostly originate from the syntactic parsing field.~\citet{muller-etal-2012-constrained} propose a method of deriving discourse dependency structures from SDRT graphs.

~\citet{lee2006complexity} points out that discourse structure is likely to be less complex than syntactic structure. The biaffine neural dependency parser by~\citet{Manning2019} achieves good performance in syntactic parsing, and in this study, we investigate empirically whether it also performs well if applied to discourse parsing. The SciDTB corpus~\citep{yang-li-2018-scidtb} is a domain-specific and manually-annotated discourse corpus with data from abstracts of the ACL anthology. It uses dependency trees to represent discourse structure. With this corpus, it is possible to develop a discourse dependency parser directly.  

\section{Background}
\subsection{Graph-based dependency parsing}
Dependency parsing generally can be implemented in transition-based approaches or graph-based approaches. In studies using graph-based approaches, dependency parsing is treated as a task of finding the highest-scored dependency tree. The arc-factored model proposed by~\citet{mcdonald-etal-2005-non} is commonly used. The score of an edge is defined as the dot product between a high-dimensional feature representation and a weight factor, and the score of a dependency tree is obtained by summing the scores of the edges in the tree. Margin Infused Relaxed Algorithm (MIRA)~\citep{crammer2003ultraconservative} is generally used to learn the weight factors. The Eisner algorithm~\citep{eisner1996three} and the Chu-Liu-Edmonds algorithm~\citep{chu1965shortest, edmonds1967optimum} can be used to find the highest-scored dependency tree from the scores of all the possible arcs. The Eisner algorithm is a dynamic programming approach and because of its restriction on the position and orientation in finding the head, it can only generate projective structures. The Chu-Liu-Edmonds algorithm aims at finding the maximum spanning tree (MST) from the scores and takes an iterative approach in removing cycles. It can produce non-projective structures.   

\citet{li-etal-2014-text}'s discourse dependency parser is an example using this approach. Since there were no discourse corpora annotated with dependency structure, an algorithm for converting RST trees to dependency trees is used. However, based on the study by~\citet{hayashi-etal-2016-empirical}, the  dependency trees converted with this algorithm are all projective, and this may be the reason why in~\citet{li-etal-2014-text}'s experiments, the accuracy is higher when the Eisner algorithm is used than when the MST algorithm is used.  

\subsection{Neural graph-based dependency parsing}
 The work by~\citet{kiperwasser-goldberg-2016-simple} addresses the problem of feature representation in developing neural dependency parsers. Instead of using feature templates, in their study, each word is represented by embedding vectors, which are fed to BiLSTM layers. The resulting BiLSTM representations are scored using a multi-layer perceptron (MLP) and trained together with the parser to learn feature representations for the specific task.

On this basis,~\citet{Manning2019} propose a biaffine dependency parser. The embedding vectors of a sequence of words are passed to a multi-layer BiLSTM network. Instead of using an MLP scorer, the output from the BiLSTM network is fed to four MLP layers, from which four vectors are produced. These vectors are passed to two biaffine layers, one deciding the most probable head for a word, and the other determining the label for the pair formed by the word and its head predicted by the previous biaffine layer. The two biaffine classifiers are trained jointly to minimize the sum of their softmax cross-entropy losses. To ensure the tree is well-formed, Tarjan's algorithm is used to remove cycles\footnote{\url{https://github.com/tdozat/Parser-v1/blob/master/lib/etc/tarjan.py}}.

\section{Experiment}
One of the major differences between syntactic parsing and discourse parsing is that syntactic parsing normally takes individual words as the basic unit, while discourse parsing treats elementary discourse units (EDU)s, which are generally phrases, as the basic unit. 

\subsection{Corpus}
In the SciDTB corpus\footnote{\url{https://github.com/PKU-TANGENT/SciDTB/tree/master/dataset}}, we extract the values of the ``text'', ``parent'' and ``relation'' fields, which represent the EDU, the gold head of the EDU, and the gold relation for this EDU and its gold head, respectively.  

The original division of the training, development and test sets of the corpus is kept. For the development and test tests, we only use data inside the ``gold'' folder. Thus, we obtain 743 documents in the training set, 154 documents in the development set and 152 documents in the test set. 

The root EDU is represented by a single symbol ``ROOT'' in the corpus. In the pre-processing step, we skip this word when doing tokenization using NLTK~\citep{bird-loper-2004-nltk}. To avoid mistaking it for a real word, we reserve a special id for this symbol when building the vocabularies of characters and words for use in the embedding layer. 

\subsection{Biaffine Discourse Dependency Parser}
For a document $\mathit{d_i}$ in the corpus, the input consists of $\mathit{m}$ EDUs $\mathit{EDU_1...EDU_{m}}$. For $\mathit{EDU_k}$ $\mathit{(1\le k \le {m})}$, which is comprised of $\mathit{n}$ words $\mathit{w_1...w_{n}}$, the embeddings of a word $\mathit{w_j}$ $\mathit{(1\le j \le {n})}$ in $\mathit{EDU_k}$ are obtained by\footnote{The $\oplus$ symbol denotes concatenation.}:  
\begin{center}
$\mathit{E_{w_j} = Emb_{w_j}\oplus CharEmb_{w_j}}$
\end{center}
Similar to~\citet{Manning2019}, we use pretrained 100-dimensional GloVe embeddings~\citep{pennington-etal-2014-glove} for $\mathit{Emb_{w_j}}$. 

To obtain $\mathit{CharEmb_{w_j}}$, we break down a word into characters and pad the words, EDUs and documents to maximum lengths to be processed efficiently. The dimension of the input to the character-level BiLSTM layer is $52$ (EDUs per document) *$46$ (words per EDU) * $45$ (characters per word).
Similar to~\citet{zhang-etal-2020-efficient}, only one BiLSTM layer is used to obtain the character-level embeddings. We set the hidden size of the character BiLSTM model to 25 and the dimension of the output to 50. 

The output character embedddings are concatenated with the 100-dimensional word vectors, thereby obtaining 150-dimensional word embeddings. Then, the embedding of $\mathit{EDU_k}$ is obtained by:
\begin{center}
$\mathit{E_{EDU_k} = E_{w_1}\oplus...\oplus E_{w_{n}}}$
\end{center}

The resulting input dimension becomes $52$ (EDUs per document) *$46$ (words per EDU) *$150$ (dimension of word vectors).  

The embeddings go through a dropout layer, and are then fed to a three-layer BiLSTM network, followed by a dropout layer. Four MLP layers are configured after this dropout layer, and each of the MLP layers is configured with the same dropout rate. The output from this step is then passed to two biaffine layers to predict the head of an EDU and the label for the arc formed by the EDU and the predicted head, respectively.

\begin{figure}[H]
\begin{center}
  \includegraphics[width=35mm,height=45mm]{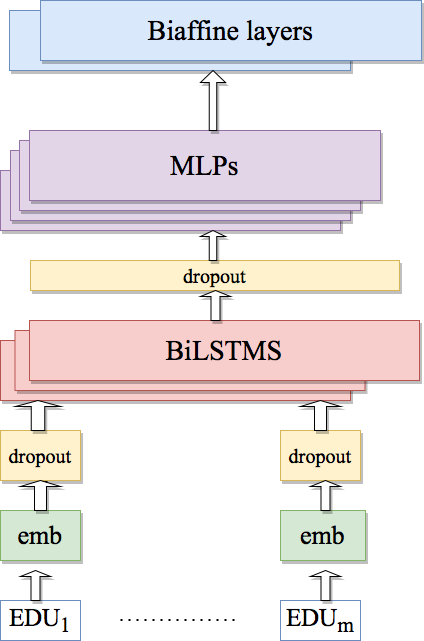}
  \caption{The architecture of the model.}
  \label{model-arch}
\end{center}
\end{figure}

The architecture of the model is shown in Figure~\ref{model-arch}.

The $52$*$52$-dimensional scores learnt from this step are processed by the Eisner algorithm and the Chu-Liu-Edmonds algorithm separately in two experiments for comparison.  

As shown in Table~\ref{biaffine_para_config}, the hyperparameters are set following~\citet{Manning2019}.
\begin{table}[H]
\begin{adjustbox}{width=0.5\columnwidth, height=20mm, center}
\begin{tabular}{l|l}
\hline
\textbf{Hyperparameter} & \textbf{value}  \\
\hline
Embedding size & 150 \\
Embedding layer dropout & 0.33\\
BiLSTM hidden size & 400 \\
BiLSTM layers  &  3   \\
BiLSTM dropout & 0.33 \\
Arc\_MLP output dimension & 500  \\
Relation\_MLP output dimension & 100 \\
MLP dropout & 0.33  \\
\hline
\textbf{Data\_info} & \textbf{value} \\
\hline
word\_vocab size & 7836 \\
char\_vocab size & 79\\
headset size & 45 \\
relationset size &27 \\
\hline
\end{tabular}
\end{adjustbox}
\caption{Hyperparameter configuration and data information.}
\label{biaffine_para_config}
\end{table}

The gold head and gold relation labels obtained from the corpus are padded with zeros to match the dimension of the input. 

We use the cross-entropy loss function to calculate the loss of predicting the heads and the loss of predicting the relations, and the sum of the two is the total loss to be minimized. We set the learning rate to 0.001, and use the Adam optimizer (lr=0.005, betas=(0.9, 0.9), weight\_decay= 1e-5). The batch size is set to 15. 

We use the PyTorch framework~\citep{NEURIPS2019_9015} (version: 1.8.0+cu111) to build the model, which is trained on RTX2060 Super GPU. The best performance is decided based on the performance on the development set. 

The evaluation metrics are the unlabeled attachment score (UAS), which measures the accuracy of predicted heads, and the labeled attachment score (LAS), which measures the accuracy of predicting the relations between the EDUs and their heads when the heads have been correctly predicted. As our basic unit is EDU, punctuation is not an issue in the evaluation. However, as the PyTorch framework requires the output classes to be in the range of [0, number\_of\_classes), the model cannot produce head ``-1'' for the root EDU. Therefore, we do not count the root EDU in the evaluation. Moreover, we use the original length of each document to filter out padded zeros to avoid discarding labels encoded as zero or potentially inflating the accuracy by considering the performance on padded zeros.   

The model achieves the best performance at the 68\textsuperscript{th} epoch for the Chu-Liu-Edmonds algorithm, and the training time is 15 minutes. When the Eisner algorithm is used, the best performance is achieved at the 86\textsuperscript{th} epoch. The training takes 25 minutes. As the basic unit is EDU and the texts are short, the training time does not form a bottleneck in our experiments. 

Table~\ref{experiment_result} shows the result of the experiments. 
\begin{table}[H]
\begin{adjustbox}{width=\columnwidth,center}
\begin{tabular}{|l|l|l|l|l|}
\hline
\textbf{Experiment} & \multicolumn{2}{c|}{\textbf{dev set}} & \multicolumn{2}{c|}{\textbf{test set}}\\
\hline
      & UAS & LAS & UAS & LAS \\
\hline 
Eisner & 0.694 & 0.601 & 0.692 & \textbf{0.573} \\
\hline
Chu-Liu-Edmonds & \textbf{0.753}  & \textbf{0.616} & \textbf{0.729}  & 0.571 \\
\hline
\textbf{Baseline parsers}~\citep{yang-li-2018-scidtb}& \multicolumn{4}{c|}{\textbf{ }} \\
\hline
Graph-based & 0.607 & 0.455 & 0.576 &  0.425 \\
\hline
Two-stage transition & 0.730 & 0.577 & 0.702 &  0.545 \\
\hline
Human agreement & 0.806 & 0.627 & 0.802 &  0.622 \\
\hline
\end{tabular}
\end{adjustbox}
\caption{\label{experiment_result} Result of the experiment.}
\end{table}
The model achieves higher accuracy than any of the baseline parsers, whether transition-based or graph-based, although our model is only a first-order parser and the embeddings used in the experiments are simple. With the Chu-Liu-Edmonds algorithm, the model achieves higher accuracy than when the Eisner algorithm is used. 

Inspired by the study by \citet{ferracane-etal-2019-evaluating}, to evaluate the structure of the output, we calculate the average maximum path length and the proportion of leaf nodes of the dependency trees generated by the model on the development and test sets. The average maximum path length is obtained by dividing the sum of the longest path lengths of dependency trees predicted by the parser by the number ${\mathit{n}}$ of documents:   
\begin{center}
$\mathit{\small Avg\_max\_path\_len=\frac{\sum_{i=1}^{n}Max\_Path\_Len_{d_i}}{n}}$
\end{center}

The average proportion of leaf nodes is computed by dividing the number of leaf nodes, which are identified by finding nodes with out-degree zero and in-degree not equal to zero, by the total number of nodes in a dependency tree of $\mathit{d_i}$, and summing the proportions and dividing the sum by the total number ${n}$ of documents in the dataset. 
\begin{center}
$\mathit{Avg\_pro\_leafnode=\frac{\sum_{i=1}^{n}(C\_LeafNodes_{d_i}/C\_Nodes_{d_i})}{n}}$
\end{center}

From Table~\ref{tree-depth}, we can see that the predicted trees are similar to the gold trees. The proportion of leaf nodes does not show much variation between the Chu-Liu-Edmonds algorithm and the Eisner algorithm, but with the Chu-Liu-Edmonds algorithm, the tree path length tends to be longer.
From these two metrics, we may conclude that the model is not likely to produce ``vacuous trees''~\citep{ferracane-etal-2019-evaluating}. 

\begin{table}[H]
\begin{adjustbox}{width=0.7\columnwidth, height=13mm, center}
\begin{tabular}{|l|l|l|}
\hline
\textbf{Avg max path len} & \textbf{dev set} & \textbf{test set}\\
\hline
Eisner & 4.318 & 4.303 \\
\hline
Chu-Liu-Edmonds & 4.415 & 4.329 \\
\hline
Gold & 4.474 & 4.447 \\
\hline
\textbf{Avg proportion of leaf nodes} &  & \\
\hline
Eisner & 0.447 & 0.452 \\
\hline
Chu-Liu-Edmonds & 0.447 & 0.457 \\
\hline
Gold & 0.450 & 0.455 \\
\hline
\end{tabular}
\end{adjustbox}
\caption{\label{tree-depth}Average maximum path length and average proportion of leaf nodes.}
\end{table}

\section{Complexity of Discourse Dependency Structure}
As the SciDTB corpus allows non-projective structures, we make an investigation of the complexity of non-projectivity of the corpus in the formal framework proposed by~\citet{kuhlmann-nivre-2006-mildly}. The result is shown in Table~\ref{non-proj-metrics}.
\begin{table}[H]
\begin{adjustbox}{width=0.3\columnwidth,height=10mm, center}
\begin{tabular}{l|l}
\hline
\textbf{Property} & \textbf{value} \\
\hline
gap degree 0 & 1014\\
gap degree 1 & 35  \\
\hline
edge degree 0 & 1014  \\
edge degree 1 & 35  \\
\hline
projective & 1014 \\
non-projective & 35\\
\hline
\end{tabular}
\end{adjustbox}
\caption{\label{non-proj-metrics}
Complexity of nonprojectivity in SciDTB.}
\end{table}

Gap degree measures the number of discontinuities within a subtree while the edge degree measures the number of intervening components spanned by a single edge~\citep{kuhlmann-nivre-2006-mildly}. From Table~\ref{non-proj-metrics}, we can see that the majority of the documents in the corpus are projective and the dependency structures have gap degree at most one and edge degree at most one. This finding provides evidence for~\citet{li-etal-2014-text}'s observation that discourse dependency structures are simple and may be computationally efficient to process. 

\section{Conclusion and Future Work}
We apply the biaffine model to discourse dependency parsing, which achieves high accuracy and generates tree structures close to the gold trees. In this way, we show that the dependency framework is effective for representing discourse. 

Since the SciDTB corpus is formed by abstracts of scientific papers, the parser we develop can only handle short texts. In future work, we plan to investigate the application of the dependency framework in discourse parsing of longer texts from more languages and domains. 
\bibliography{anthology,custom}
\bibliographystyle{acl_natbib}
\end{document}